\documentclass[11pt]{article}

\usepackage[final]{acl}

\usepackage{times}
\usepackage{latexsym}

\usepackage[T1]{fontenc}

\usepackage[utf8]{inputenc}

\usepackage{microtype}

\usepackage{inconsolata}

\usepackage{graphicx}

\usepackage{amssymb}
\usepackage{amsfonts}

\usepackage{subcaption}
\usepackage{enumitem}
\usepackage{algorithm}
\usepackage{algpseudocode}
\usepackage{amsmath} %
\usepackage{xcolor}
\usepackage{listings}
\usepackage[most]{tcolorbox}
\usepackage{capt-of}
\tcbset{
  promptbox/.style={
    enhanced,
    colback=gray!3,
    colframe=gray!50,
    boxrule=0.5pt,
    arc=2pt,
    left=6pt,right=6pt,top=6pt,bottom=6pt,
    fonttitle=\small\bfseries,
    coltitle=black,
    breakable,
    toprule at break=0pt,
    bottomrule at break=0pt,
  }
}
\newtcblisting{promptlisting}[1][]{
  promptbox,
  listing engine=listings,
  listing only,
  title={#1},
  listing options={
    basicstyle=\ttfamily\small,
    columns=fullflexible,
    breaklines=true,
    breakatwhitespace=false,
    breakautoindent=false,
    breakindent=0pt,
    keepspaces=true,
    showstringspaces=false,
    escapechar=|,
    xleftmargin=0pt,
    aboveskip=0pt,
    belowskip=0pt
  }
}

\title{Where Reasoning Breaks: Logic-Aware Path Selection by Controlling Logical Connectives in LLMs Reasoning Chains}

\author{Seunghyun Park \\
  Independent Researcher\\
  \texttt{qkrpsh97@gmail.com} \\\And
  Yuanyuan Lei \\
  University of Florida\\
  Gainesville, FL, United States\\
  \texttt{yuanyuan.lei@ufl.edu} \\}

\begin{document}
\maketitle
\begin{abstract}
While LLMs demonstrate impressive reasoning capabilities, they remain fragile in multi-step logic deduction, where a single transition error can propagate through the entire reasoning chain, leading to unstable performance. In this work, we identify \textit{logical connectives} as primary points of this structural fragility. Through empirical analysis, we show that logical connective tokens function as high entropy forking points, at which models frequently struggle to determine the correct logical direction. Motivated by this observation, we hypothesize that intervening in logical connective selection can guide LLMs towards the correct logical direction, thereby improving the overall reasoning chain. To validate this hypothesis, we propose a multi-layered framework that intervenes specifically at these logic-critical junctions in the reasoning process. Specifically, we introduce (1) \textit{Gradient-based Logical Steering} to guide LLMs internal representations towards valid reasoning subspaces, (2) \textit{Localized Branching} to resolve ambiguity via targeted look-ahead search, and (3) \textit{Targeted Transition Preference Optimization}, a surgical reinforcement learning objective that selectively optimizes single-token preferences at logical pivots. Crucially, by concentrating intervention solely on logic-critical transitions, our framework achieves a favorable accuracy--efficiency trade-off compared to global inference time scaling methods like beam search and self-consistency\footnote{The code link is: \url{https://github.com/lei-nlp-lab/reasoning_fragility_acl_2026}}.

\end{abstract}

\section{Introduction}
Logical reasoning \citep{zhang2025empoweringllmslogical, zhang2025logicalreasoninglarge} serves as a cornerstone of general intelligence, enabling large language models (LLMs) to tackle rigorous domains such as complex decision-making, program synthesis, or mathematics \citep{wei2024llmmastermindsurvey, zhang2025systemsystemsurvey, he2025codeiocondensingreasoning, luo2025codethinkthink}. Unlike open-ended generation, logical reasoning requires the model to construct a coherent chain of thought, where valid conclusions are derived from premises through a structured, multi-step deductive process \citep{Wei2022ChainOT, lightman2023let, liu2025safe, mcginness2024automated}. In this strict framework, the integrity of the sequence is paramount: a single flaw in an intermediate step can propagate into a complete failure of the reasoning chain. Therefore, ensuring the precision of each transition within the reasoning process is critical for reliable model performance.

\begin{figure*}[ht]
  \centering
  \includegraphics[width=0.9\linewidth]{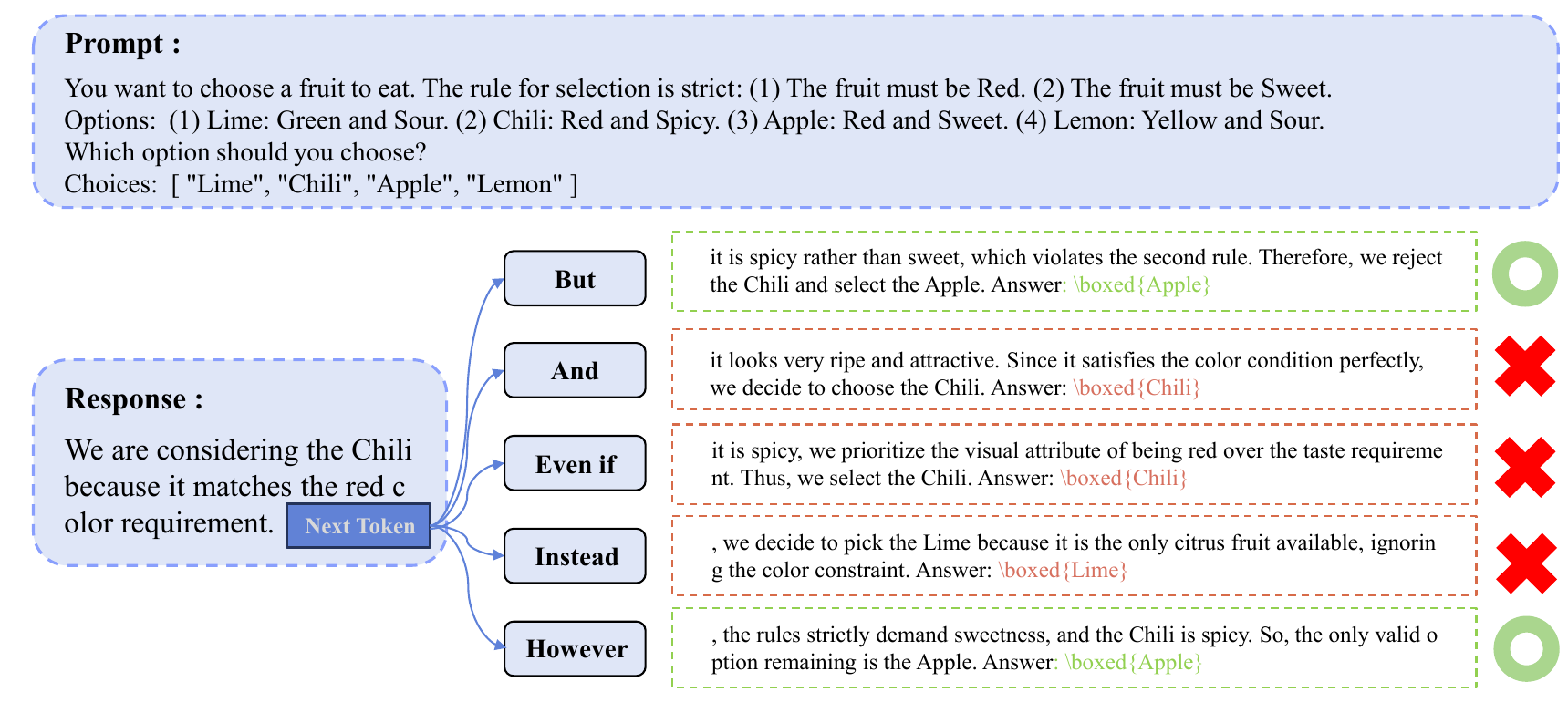}
\caption{Example showing the fragility of reasoning at logical connective pivots: replacing only the logical connective at a single transition flips the final answer.}
\label{fig:intro_example}
\end{figure*}

We observe that \textit{logical connectives} serve as critical points of fragility in reasoning. Logical connectives, such as \textit{therefore}, \textit{however}, and \textit{but}, function as linguistic pivots that explicitly direct the logical links between successive reasoning steps. Throughout this paper, we use the term \textit{logical connectives} to refer to explicit discourse markers that signal logical relations between reasoning steps, following the discourse relation framework of \citet{robaldo2008penn}
. These tokens act as directional signals: a single connective chosen at the wrong step can redirect the reasoning chain and ultimately determine whether the deduction is valid or incorrect. This high-leverage role is illustrated in Figure~\ref{fig:intro_example}, where replacing only the logical connective is sufficient to flip the final conclusion, revealing that valid reasoning often hinges on these specific transition points.

\begin{figure*}[t]
  \centering

  \begin{subfigure}[t]{0.49\linewidth}
    \centering
    \includegraphics[width=\linewidth,height=3.0cm,keepaspectratio]{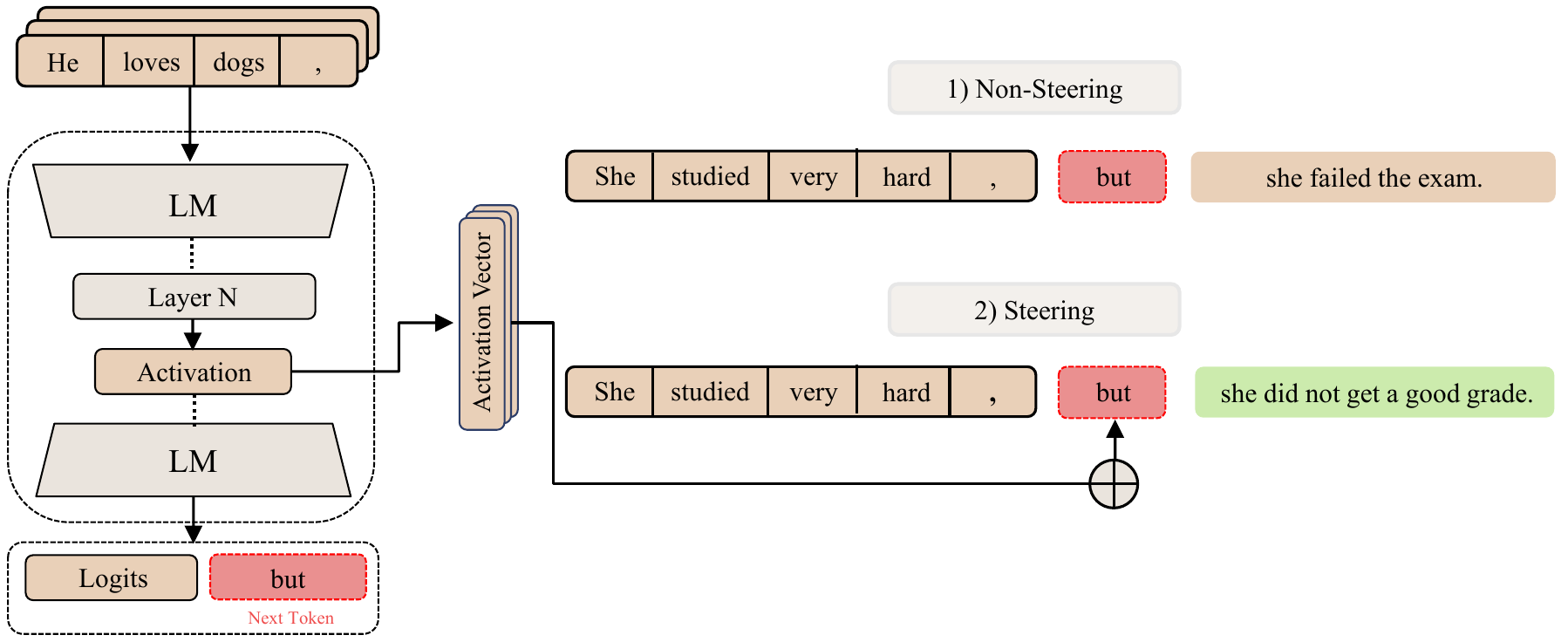}
    \caption{Gradient-based Logical Steering}
    \label{fig:methods_steering}
  \end{subfigure}\hfill
  \begin{subfigure}[t]{0.49\linewidth}
    \centering
    \includegraphics[width=\linewidth,height=3.0cm,keepaspectratio]{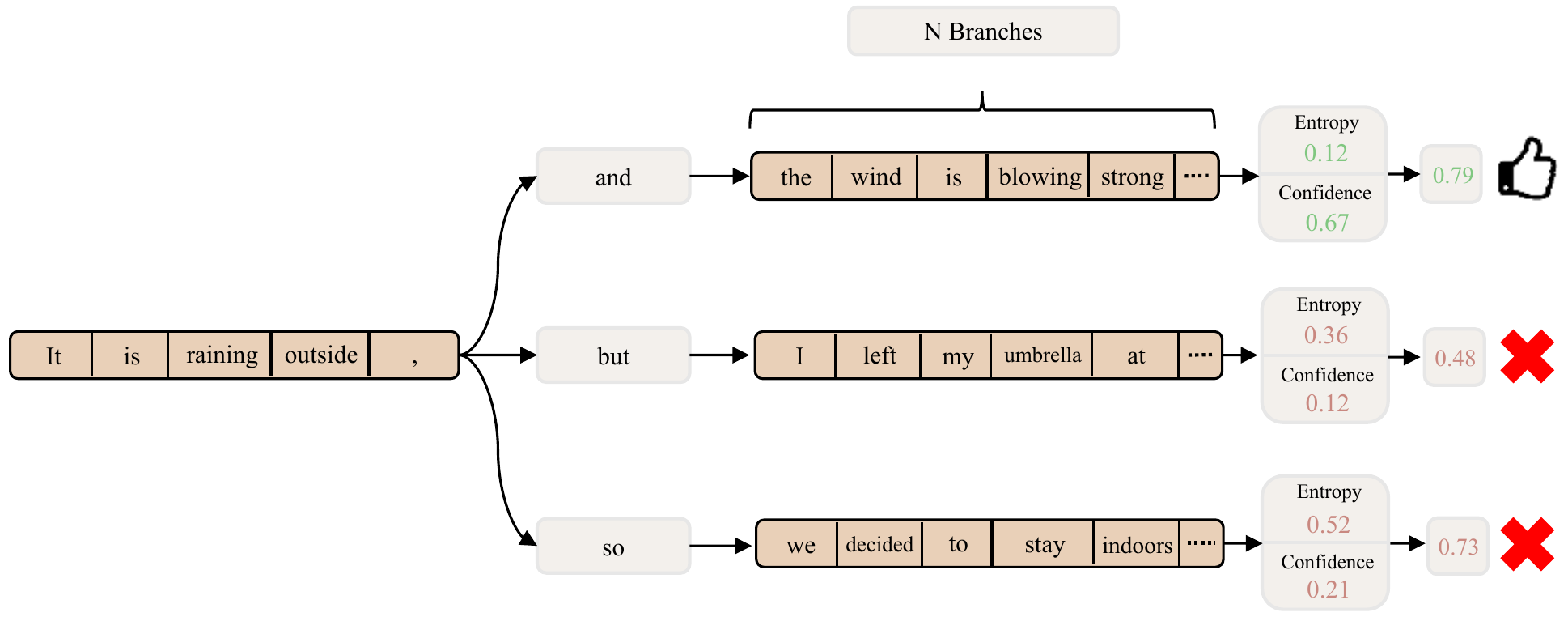}
    \caption{Localized Branching}
    \label{fig:methods_branching}
  \end{subfigure}

  \begin{subfigure}[t]{\linewidth}
    \centering
    {\includegraphics[width=0.72\linewidth,height=3.0cm,keepaspectratio]{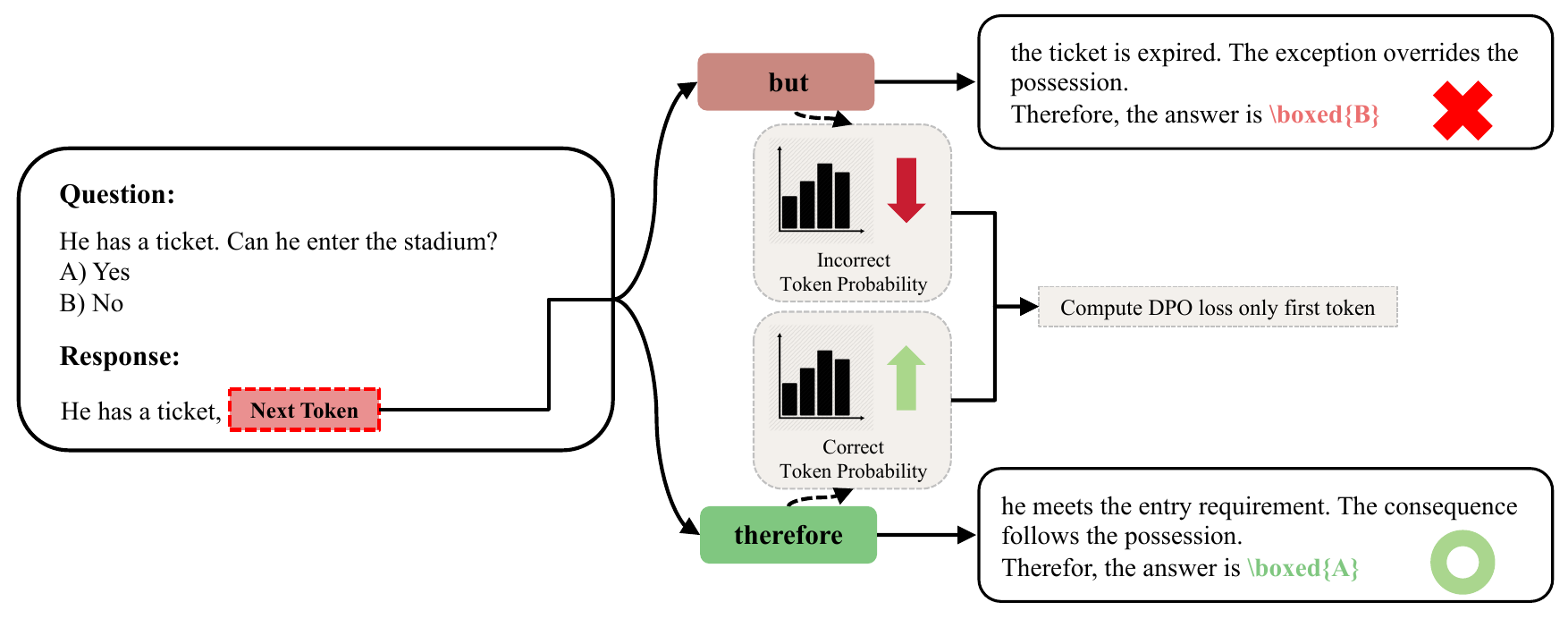}}
    \caption{Targeted Transition Preference Optimization (TTPO)}
    \label{fig:methods_ttpo}
  \end{subfigure}

  \caption{Connective centric methods across three stages. (a) Steering provides training-free activation intervention at connective junctions. (b) Branching provides compute localized test time search by scoring short continuations for alternative connectives. (c) TTPO provides minimal tuning by optimizing only the single token preference at connective pivots, correcting the underlying distribution that causes brittle transitions. }
  \label{fig:methods}
\end{figure*}

To further validate this observation, we conduct a diagnostic analysis to uncover the fragility of reasoning process in LLMs (Section~\ref{sec:preliminary}). Our analysis reveals that logical connectives are pivot points of reasoning fragility and function as critical decision junctions. Firstly, token-level entropy analysis shows that uncertainty is predominantly concentrated at connective positions. Secondly, inspection of the top candidates in the model’s distribution indicates that the model's decision primarily focuses on which connective to use, rather than refining the surrounding content. Thirdly, substituting connective tokens alone produces statistically significant changes in the final reasoning performance, with such connective perturbations derailing correct chains nearly twice as often as perturbations at other high-entropy positions. Furthermore, analysis at the level of logical relations reveals that successful reasoning repairs occur when connective perturbations drive transitions between different relation types, rather than token substitutions within the same relation category, which confirms that connective choices direct the logical trajectory of the reasoning chain. Taken together, these findings suggest that logical connectives operate as ambiguous forking points where the model struggles to determine the next logical direction, and small mistakes at these junctures can propagate and ultimately degrade the entire reasoning chain.

Motivated by the above observation and analysis, we arrive at the following core hypothesis: \textit{Optimizing logical connectives as the key decision points will guide the model to select a valid logic-aware reasoning path, thereby leading to correct reasoning answer and improving reasoning performance.}

To substantiate this hypothesis, we propose a multi-layered framework that intervenes directly at these logic-critical junctions and provides focused control across three complementary levels. To elaborate: (i) at the \textit{activation level}, we design a gradient-based steering mechanism to steer LLMs internal representations toward valid logic directions, enabling recovery from failing reasoning chains without modifying model weights \citep{turner2308steering, rimsky2024steering}, (ii) at the \textit{inference level}, we introduce localized branching, a look-ahead strategy that computes an ambiguity score at connective positions and selectively explores only the most promising logic-guided paths \citep{zhao2024lookahead, chen2023accelerating}, (iii) at the \textit{training level}, we propose \textit{Targeted Transition Preference Optimization}, a surgical reinforcement learning objective that optimizes single-token preferences at logical pivots to refine the model's distribution. Together, these three strategies guide LLMs towards an optimal logic-aware reasoning path by controlling the logic transition points where reasoning failures most often originate.

The experiments on five logical reasoning benchmarks and four LLMs demonstrate that by shifting the focus from the entire sequence to these high leverage transitions, our approach achieves performance gains while maintaining the efficiency of greedy decoding. Our main contributions are:
\begin{itemize}
    \item We empirically diagnose logical connectives as the primary points of fragility in reasoning
    \item We introduce a multi-layered framework to guide LLMs towards a logic-aware reasoning path by intervening directly at logic junctions
\end{itemize}

\section{Related Work}

\paragraph{Logical Reasoning}
Logical reasoning \citep{zhang2025logicalreasoninglarge, yang2023logical} in NLP generally refers to the ability to derive valid conclusions from a given set of premises. While LLMs have demonstrated emergent capabilities in deductive reasoning \citep{creswell2022selection, wei2022chain}, they often struggle with \textit{compositionality}, the ability to maintain logical consistency across long, multi-step chains.
Previous works \citep{creswell2022selection, wei2022chain} have attempted to bridge this gap by decomposing reasoning into explicit 'Selection-Inference' steps or by integrating neuro-symbolic modules \citep{creswell2022faithful, pan2023logic} that translate natural language into formal logic representations.
However, these approaches often require rigid formalisms or external solvers. In contrast, our work focuses on the \textit{intrinsic linguistic mechanism} of reasoning within the model. We posit that the failures in multi-step deduction often stem from imprecise transitions at the token level, specifically at logical connectives, which serve as the natural language operators for deduction.

\paragraph{Inference Time Intervention}
Test-time scaling methods improve reasoning by allocating extra compute via global search or sampling (e.g., Self-Consistency, ToT) \citep{lightman2023let, chen2025towards, wang2022self, yao2023tree}, but they are often compute-inefficient because they scale over entire trajectories.
We instead localize intervention to \textit{connective pivots}: (i) activation-level steering nudges hidden states without weight updates \citep{panickssery2312steering, turner2308steering}, and (ii) pivot triggered lookahead branching explores only when connective ambiguity is detected \citep{zhao2024lookahead, chen2023accelerating}.
This yields a targeted alternative to global scaling while preserving near greedy decoding behavior.


\paragraph{Critical Tokens and Token-Level Analysis}
Recent studies show that uncertainty is often dominated by a small set of high-entropy tokens that act as branching points, and that editing such tokens can disproportionately change the final answer \citep{wang2025beyond, zur2025language, bogdan2025thought, lin2024critical}.
We adopt this token level perspective but focus on a linguistically grounded subset: logical connectives.
Using entropy statistics on CoT traces, we find that high entropy events concentrate on connective positions, and we intervene directly at these junctions during generation via both inference time mechanisms and a targeted training objective.

\section{Preliminary}
\label{sec:preliminary}

In this section, we empirically validate our core hypothesis: \textit{logical connectives act as the primary decision-making pivots in CoT reasoning, yet they represent points of fragility in current LLMs.} We conduct our pilot analysis primarily on ZebraLogic \citep{lin2025zebralogic}, and additionally use the deductive subset of BIG-Bench Hard (BBH) \citep{suzgun2023challenging} as a controlled testbed for single token replacement experiments.

\subsection{High Entropy Rate of Logical Connectives}

\begin{figure}[ht]
\centering
\includegraphics[width=1.0\linewidth]{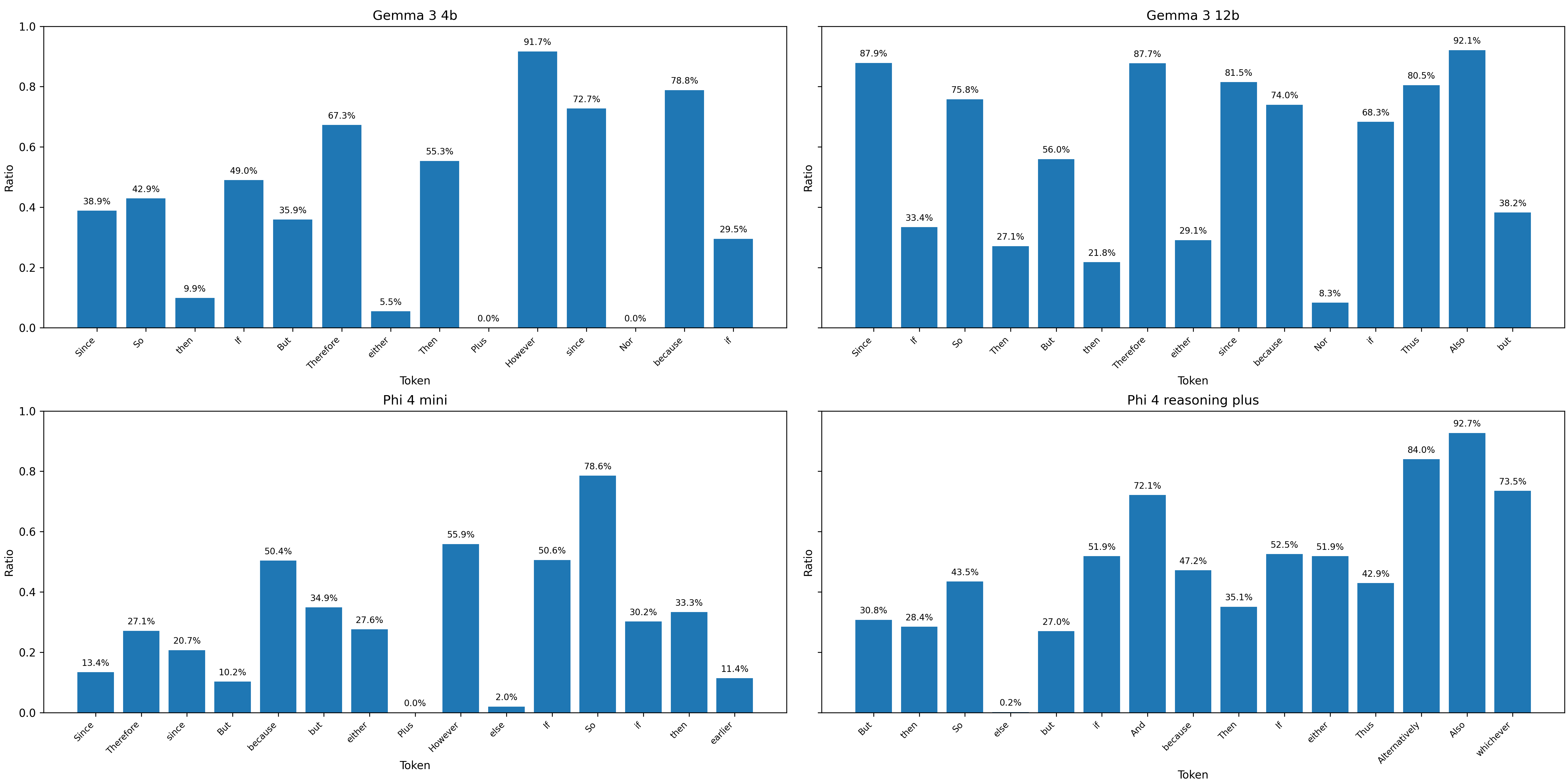}
\caption{Analysis of token entropy at logical connective positions. The proportion of connectives exceeding the entropy threshold $\tau=1.0$.}
\label{fig:logical_token_ratio}
\end{figure}

To quantify the intrinsic uncertainty associated with logical transitions, we measure the \textbf{High Entropy Rate ($R_{HE}$)} specifically for the set of logical connectives $\mathcal{S}_{l}$ (Appendix~\ref{sec:logical_connectives}) on the ZebraLogic dataset. This metric represents the probability that the model encounters high uncertainty when generating a connective. Following our experimental setup, we define $R_{HE}$ as:

\begin{equation}
R_{HE}(\mathcal{S}_{l}) = \frac{\sum_{t} \mathbb{I}(w_t \in \mathcal{S}_{l} \wedge H_t > \tau)}{\sum_{t} \mathbb{I}(w_t \in \mathcal{S}_{l})}
\end{equation}

where $\mathbb{I}(\cdot)$ is the indicator function and $\tau = 1.0$ is the entropy threshold. 

As illustrated in Figure~\ref{fig:logical_token_ratio}, our empirical analysis reveals that a substantial proportion of logical connectives are generated under high entropy conditions. To ensure this finding is not an artifact of a specific threshold, we conducted two supplementary analyses (detailed in Appendix~\ref{sec:entropy_analysis}). First, a threshold free quantile enrichment analysis shows that connectives are over-represented in every high-entropy tail by a factor of 2.0--5.8$\times$, despite constituting only $\sim$4--7\% of all generated tokens. Second, a comparison across token categories over a broad $\tau$ sweep confirms that logical connectives consistently exhibit the highest $R_{HE}$ among all categories at every threshold tested.

\subsection{Logical Ambiguity in Top-K Candidates}

\begin{figure}[ht]
\centering
\includegraphics[width=1.0\linewidth]{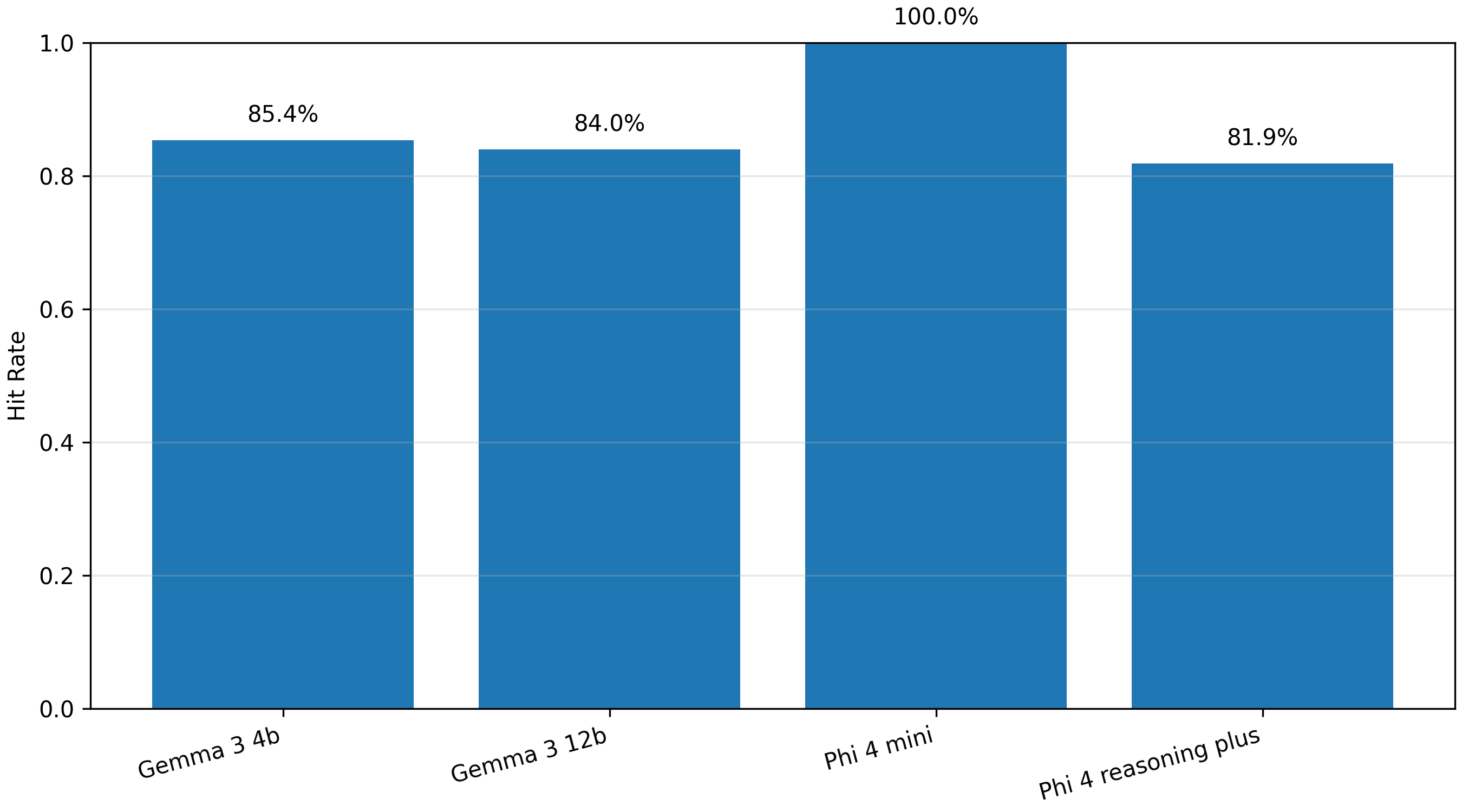}
\caption{Logical connective presence in the Top-$5$ candidate set across models.}
\label{fig:topk_hit_rate}
\end{figure}

To determine if high entropy reflects mere stylistic variation or genuine logical ambiguity, we inspect the Top-$5$ candidate distribution at these junctions, again on ZebraLogic. We find that in the majority of high entropy cases, the Top-$5$ set contains at least one alternative logical connective. 

This indicates that these positions are \textit{forking points}, the model's internal representations are simultaneously entertaining multiple, divergent reasoning trajectories.

\subsection{Disproportionate Causal Leverage of Logical Connectives}
\label{sec:causal_leverage}

To evaluate the causal leverage of logical connectives, we perform single-token perturbation experiments on both ZebraLogic and the BBH deductive subset. Specifically, we replace exactly one greedily selected connective with a random alternative from $\mathcal{S}_{l}$ and observe the impact on the final reasoning outcome.

\begin{table}[ht]
\centering
\small
\begin{tabular}{llc}
\hline
\textbf{Original $\to$ Modified} & \textbf{Label} & \textbf{Rate (\%)} \\
\hline
Correct $\to$ Correct & C $\to$ C & 23.4 \\
Correct $\to$ Incorrect & C $\to$ I & 16.3 \\
Incorrect $\to$ Incorrect & I $\to$ I & 50.5 \\
Incorrect $\to$ Correct & I $\to$ C & 9.8 \\
\hline
\end{tabular}
\caption{Distribution of answer correctness under a single random logical connective replacement on ZebraLogic.}
\label{tab:flip_rate}
\end{table}

As shown in Table~\ref{tab:flip_rate}, a single random token change at a logical junction causes a significant divergence in the final answer. To properly interpret this causal impact, we measure two conditional metrics. First, we observe a \textit{conditional fragility} of 41.1\% among originally correct chains ($\frac{C \to I}{C \to C + C \to I}$), meaning nearly half of the correct reasoning trajectories are completely derailed by changing just one connective. 

Conversely, we observe a \textit{conditional repair rate} of 16.2\% among originally incorrect chains ($\frac{I \to C}{I \to I + I \to C}$). This rate represents a strict lower bound on repairability, as it reflects the outcome of \textit{random} connective selection; our targeted methods (Section~\ref{sec:method}), which guide the model to select connectives based on contextual signals, are designed to operate substantially above this baseline. The asymmetry between high fragility (41.1\%) and low random repair (16.2\%) reveals that the space of trajectory-derailing connectives is much larger than the space of trajectory-correcting ones, directly motivating the need for principled connective selection at these high-leverage pivots.

\begin{table}[ht]
\centering
\small
\begin{tabular}{lc}
\hline
\textbf{Single-Token Replacement Type} & \textbf{C $\to$ I Rate (\%)} \\
\hline
Connective $\to$ Random connective & 22.8 \\
Connective $\to$ Same-class connective & 21.7 \\
Non-connective $\to$ Random token & 13.0 \\
\hline
\end{tabular}
\caption{Controlled single-token replacement on the BBH dataset. Perturbing a connective derails the correct reasoning chain nearly twice as often as perturbing a general high-entropy token.}
\label{tab:controlled_replacement}
\end{table}

Crucially, we prove that this disproportionate leverage is unique to logical connectives, rather than a generic property of any uncertain position. Specifically, we conduct a controlled single-token replacement study on the deductive subset of the BBH dataset (Table~\ref{tab:controlled_replacement}), and compare the impact of perturbing a connective against perturbing a generic non-connective token drawn from the high-entropy tail. We found that perturbing a connective derails a correct chain ($C \to I$) at a rate of 22.8\%, whereas perturbing a non-connective high-entropy token disrupts the chain only 13.0\% of the time. This occurs despite logical connectives constituting only a small fraction ($\sim$4--7\%) of all generated tokens. Furthermore, replacing a connective with another connective from the \textit{same} logical class (e.g., ``but'' to ``however'') still yields a high 21.7\% failure rate, comparable to random cross-class replacements. This disproportionate per-token impact confirms that connective positions are structurally brittle pivot points. They do not merely represent unstable decoding steps, but act as high-leverage operators that structurally amplify local perturbations into downstream trajectory divergence.

\subsection{Logical Relation Level Analysis of Connective Repairs}
\label{sec:relation_shifts}

The previous subsections establish that connective positions exhibit disproportionate causal leverage over reasoning outcomes. A natural follow-up question is whether the repairs observed at these pivots stem from genuine shifts in logical relations, or merely from stylistic redistribution of near-synonyms (e.g., swapping ``but'' for ``however'').

To address this, we analyzed the transition patterns of successful repairs at connective pivots on the BBH deductive subset using Gemma-3-4b-it. Our transition-level data reveals that the actual repairs driving performance improvements are highly concentrated in specific, semantically meaningful \textit{cross-class} relation transitions, rather than stylistic within-class swaps.

\begin{table}[ht]
\centering
\small
\begin{tabular}{llc}
\hline
\textbf{Source Class} & \textbf{Target Class} & \textbf{Repair Rate (\%)} \\
\hline
Causal & Instantiation & 38.6 \\
Causal & Analogy & 32.9 \\
Causal & Condition & 27.2 \\
Causal & Contrast & 26.9 \\
\hline
\end{tabular}
\caption{Top logical relation transitions driving reasoning repair ($I \to C$) on the BBH deductive subset (Gemma-3-4b-it). Repair Rate denotes the conditional probability of repair given the specific source$\to$target transition ($\frac{I \to C}{I \to I + I \to C}$ within each transition type). The highest-yield repairs involve redirecting a premature causal commitment to different logical operations.}
\label{tab:relation_shifts}
\end{table}

As shown in Table~\ref{tab:relation_shifts}, the most effective repairs involve redirecting the model from an incorrect \textit{causal} conclusion (e.g., ``therefore'', ``hence'') to fundamentally different logical operations. Specifically, 61.0\% of all cross-class $I \to C$ repairs originate from causal connectives. The highest-yield repairs involve shifting from a causal commitment to an \textit{instantiation} (38.6\% conditional repair rate). 

These transitions are not stylistic swaps. Replacing ``therefore'' with ``for example'' or ``unless'' fundamentally alters the logical operation from a deductive commitment to an illustration, exception, or qualification. This redirection forces the entire downstream chain into a different trajectory. This transition analysis directly confirms that correcting a reasoning chain at these pivots requires context-dependent shifts in logical relations—validating our core hypothesis that targeted intervention at connective positions fundamentally guides the model toward valid logical paths.

\section{Method}
\label{sec:method}

\subsection{Gradient-Based Logical Steering}
\label{sec:grad_steering}

To accurately capture the latent representation responsible for logical connectives, we design a gradient based steering method. While prior steering approaches \citep{rimsky2024steering, turner2308steering} often rely on averaging activation states, such methods risk capturing context specific semantic noise rather than the functional role of the connective. In contrast, our approach utilizes the gradient of the target probability, which isolates the causal direction in the activation space that maximizes the likelihood of the correct logical connective.

\paragraph{Vector Extraction}
Let $\mathcal{D}$ (Appendix~\ref{sec:data_construction}) be a dataset containing prompts paired with reference logical connectives. For a given input sequence $x$, let $w_c$ denote the correct logical connective token at index $t$. We focus on the hidden state $\mathbf{h}_t^l \in \mathbb{R}^d$ from a target layer $l$ immediately preceding $w_c$. We compute the gradient of the log probability of $w_c$ with respect to $\mathbf{h}_t^l$:

\begin{equation}
    \mathbf{g}^{(i)} = \nabla_{\mathbf{h}_t^l} \log P(w_c \mid \mathbf{h}_t^l; x_{<t})
\end{equation}

Here, $\mathbf{g}^{(i)}$ represents the direction in the representation space that locally pushes the model to generate the logical token. We aggregate these gradients over $N$ samples from OpenThoughts \citep{guha2025openthoughts} to compute the global steering vector $\mathbf{v}_{steer}$. The vector is obtained by averaging the normalized gradients:

\begin{equation}
    \mathbf{v}_{\text{steer}}
    = \frac{1}{N} \sum_{i=1}^{N} \text{Normalize}\!\left(\mathbf{g}^{(i)}\right)
\end{equation}

\paragraph{Inference Time Intervention}
During inference, we inject this extracted feature back into the model to guide reasoning when encountering logical connectives. For a new input at time step $t$ and layer $l$, the original hidden state $\mathbf{h}_t^l$ is modified by adding the steering vector scaled by a coefficient $\alpha$:

\begin{equation}
    \tilde{\mathbf{h}}_t^l = \mathbf{h}_t^l + \alpha \cdot \mathbf{v}_{steer}
\end{equation}

This operation effectively shifts the activation state towards the logical reasoning subspace without altering the model's weights. By tuning $\alpha$, we can control the strength of the logical guidance, ensuring the model remains attentive to transitions even in greedy decoding.

\subsection{Logical Connective Branching}
\label{sec:branching}

While the steering method (Section~\ref{sec:grad_steering}) implicitly guides the model within the activation space, our branching strategy intervenes explicitly in the decoding space to resolve ambiguity at critical junctions. As identified in our analysis, logical connectives often serve as high entropy `forking points'. When multiple plausible connectives appear at a pivot, we expand $K$ candidate continuations and select among the branched candidates using both signals: (i) an entropy based uncertainty estimate and (ii) a confidence score. This idea is inspired by the general principle from \citep{fu2025deep}, which ranks the branch with entropy and confidence score. To navigate these bifurcations, we propose a lookahead based selection mechanism that prioritizes the reasoning path with the highest model certainty.

\paragraph{Trigger Mechanism and Branching}
To efficiently detect intervention points without computational overhead, we define a set of target logical connectives $\mathcal{S}_{l}$ (Appendix~\ref{sec:logical_connectives}). During the greedy decoding process at step $t$, we apply a two-stage trigger criterion: branching is activated only when (i) the top-1 candidate token $w_{top1}$ belongs to $\mathcal{S}_{l}$, and (ii) at least two candidates from $\mathcal{S}_{l}$ appear within the top-$K$ predictions. This ensures that branching occurs only at genuine decision points where the model exhibits structural uncertainty, avoiding unnecessary lookahead at low-ambiguity steps.

When triggered, instead of committing to $w_{top1}$, we filter the top-$K$ candidates for tokens present in $\mathcal{S}_{l}$, forming a candidate set $\mathcal{C} = \{c^{(1)}, \dots, c^{(m)}\}$ where $m \le K$.

\paragraph{Lookahead Evaluation}
For each candidate connective $c^{(k)}$, we perform a lookahead generation of length $L$ to obtain a continuation trajectory $\tau^{(k)} = (y_1, \dots, y_L)$. We evaluate the quality of each trajectory using two complementary metrics:

\paragraph{Trajectory Entropy ($H$)}
This measures the model's uncertainty regarding the \textit{path} generated after the connective. Lower entropy implies a clearer reasoning chain.
\begin{equation}
    H(\tau^{(k)}) = \frac{1}{L} \sum_{j=1}^{L} \mathcal{H}\left( P(\cdot \mid x, c^{(k)}, \tau^{(k)}_{<j}) \right)
\end{equation}
where $\mathcal{H}$ denotes the entropy of the next token distribution.

\paragraph{Sequence Confidence ($S$)}
Entropy measures the spread of the distribution, but not the correctness. To ensure the generated path is high probability, we compute the length normalized log probability of the sequence:
\begin{equation}
    S(\tau^{(k)}) = \frac{1}{L} \sum_{j=1}^{L} \log P(y_j \mid x, c^{(k)}, \tau^{(k)}_{<j})
\end{equation}

\paragraph{Selection Strategy}
Directly combining $H(\tau^{(k)})$ and $S(\tau^{(k)})$ can be scale sensitive.
We therefore normalize each signal \emph{across the candidate set at the current pivot}.
Let $\{H_k\}_{k=1}^{m}$ and $\{S_k\}_{k=1}^{m}$ denote the entropy and confidence values for candidates in $\mathcal{C}$.
We compute
\begin{equation}
\tilde{H}_k = \frac{H_k - \mu_H}{\sigma_H+\epsilon},
\qquad
\tilde{S}_k = \frac{S_k - \mu_S}{\sigma_S+\epsilon},
\end{equation}
where $(\mu_H,\sigma_H)$ and $(\mu_S,\sigma_S)$ are the mean and standard deviation over $\mathcal{C}$, and $\epsilon$ is a small constant.
We then select the branch using a joint score:
\begin{equation}
c^* = \operatorname{argmin}_{c^{(k)} \in \mathcal{C}}
\left(\tilde{H}_k - \tilde{S}_k\right).
\end{equation}

This lookahead ensures that the chosen connective leads to a coherent and confident reasoning trajectory, mitigating the error propagation typical of standard greedy decoding.

\subsection{Targeted Transition Preference Optimization (TTPO)}
\label{sec:token_optimization}

While inference time strategies (Sections \ref{sec:grad_steering} and \ref{sec:branching}) effectively guide the model, they do not rectify the underlying probability distribution that causes logical errors. Traditional alignment methods like RLHF or DPO optimize preferences over entire response sequences. However, training on full sequences to correct specific local transitions is computationally expensive and may introduce gradients that degrade general language modeling capabilities on non-critical tokens.

To address this, we propose \textbf{TTPO}. This method adapts the DPO objective to focus exclusively on the single token decision step at logical branching points, serving as a highly efficient, surgical fine-tuning stage.

\paragraph{Objective Formulation}
Let $\mathcal{D}=\{(x,w_c,w_r)\}$ (Appendix~\ref{sec:data_construction}) be a dataset of triplets, where $x$ is the context right before a logical transition, $w_c$ is the chosen connective, and $w_r$ is the rejected one.
To focus on the transition step, we define a per token log ratio score
$s_\theta(x,w)=\log \pi_\theta(w\mid x)-\log \pi_{\text{ref}}(w\mid x)$.
Then TTPO maximizes the margin between $w_c$ and $w_r$ only at this step:

\vspace{-10pt}

\begin{equation}
\Delta_\theta(x,w_c,w_r)= s_\theta(x,w_c)-s_\theta(x,w_r).
\label{eq:ttpo_delta}
\end{equation}

\vspace{-15pt}

\begin{equation}
\mathcal{L}_{\text{TTPO}}
= - \mathbb{E}_{\mathcal{D}}
\big[\log \sigma(\beta\,\Delta_\theta(x,w_c,w_r))\big].
\label{eq:ttpo_loss}
\end{equation}

where $\pi_\theta$ is the policy model, $\pi_{\text{ref}}$ is the frozen reference model, and $\beta$ is a hyperparameter controlling the deviation penalty.

\paragraph{Surgical Gradient Updates}
The key advantage of TTPO lies in its \textit{gradient sparsity}. During training, we perform a forward pass only up to the logical connective position. The loss is computed solely based on the logits of $w_c$ and $w_r$, meaning gradients are backpropagated only from this specific timestep.

\begin{table*}[!t]
\centering
\scriptsize
\setlength{\tabcolsep}{3.2pt}
\renewcommand{\arraystretch}{1.12}

\begin{tabular}{lccccc||ccccc}
\hline
& \multicolumn{5}{c||}{Gemma-3-4b-it}
& \multicolumn{5}{c}{Phi-4-mini-instruct (4B)} \\
\cline{2-11}
& ZebraLogic & BBH (Ded.) & RuleBERT & LogiQA 2.0 & ProntoQA
& ZebraLogic & BBH (Ded.) & RuleBERT & LogiQA 2.0 & ProntoQA \\
\hline
\hline
Greedy (BL)         & 38.8 & 75.3 & 60.3 & 55.2 & 90.0  & 38.8 & 67.3 & 50.3 & 57.7 & 93.6 \\
Beam Search (BL)    & 33.4 & \textbf{77.0} & \textbf{67.7} & 54.0 & \textbf{90.6}
                    & 31.8 & \textbf{73.8} & 51.7 & 59.0 & 93.2 \\
Steering            & 39.0 & 75.5 & 62.0 & 55.5 & 90.2  & \textbf{39.6} & 68.8 & 50.8 & 58.3 & 94.2 \\
Branching           & \textbf{42.0} & 74.7 & 63.4 & \textbf{56.8} & 90.2
                    & \textbf{39.6} & 69.1 & \textbf{51.9} & 57.0 & 94.8 \\
TTPO                & 40.4 & 75.9 & 60.8 & 56.0 & 90.4  & 38.6 & 67.2 & 50.7 & \textbf{59.4} & \textbf{96.6} \\
\hline
& \multicolumn{5}{c||}{Gemma-3-12b-it}
& \multicolumn{5}{c}{Phi-4-reasoning-plus (13B)} \\
\cline{2-11}
& ZebraLogic & BBH (Ded.) & RuleBERT & LogiQA 2.0 & ProntoQA
& ZebraLogic & BBH (Ded.) & RuleBERT & LogiQA 2.0 & ProntoQA \\
\hline
Greedy (BL)         & 53.2 & 87.5 & 58.2 & 59.2 & 99.0  & \textbf{60.8} & 70.2 & 41.6 & 47.8 & \textbf{97.8} \\
Beam Search (BL)    & 51.8 & \textbf{89.2} & 57.5 & \textbf{60.1} & \textbf{99.2}
                    & 57.8 & 91.3 & \textbf{47.4} & \textbf{49.8} & 94.2 \\
Steering            & 53.2 & 87.8 & \textbf{59.0} & 59.4 & 99.0  & 60.2 & 88.7 & 41.4 & 47.0 & \textbf{97.8} \\
Branching           & 52.4 & 87.4 & 57.0 & 58.5 & 98.8  & 60.4 & \textbf{92.5} & 43.2 & 43.6 & 96.0 \\
TTPO                & \textbf{55.6} & 86.4 & 57.3 & 59.5 & \textbf{99.2}
                    & 58.4 & 86.4 & 43.7 & 45.7 & 97.6 \\
\hline
\end{tabular}

\caption{Overall results of steering, branching, and TTPO.}
\label{tab:main_results}
\end{table*}

\begin{table*}[t]
\centering
\scriptsize
\setlength{\tabcolsep}{3.6pt}
\renewcommand{\arraystretch}{1.12}

\begin{tabular}{lccccc||ccccc}
\hline
 & \multicolumn{5}{c||}{Gemma-3-4b-it} & \multicolumn{5}{c}{Phi-4-mini-instruct (4B)} \\
\cline{2-11}
      & ZebraLogic & BBH (Ded.) & RuleBERT & LogiQA 2.0 & ProntoQA
      & ZebraLogic & BBH (Ded.) & RuleBERT & LogiQA 2.0 & ProntoQA \\
\hline
\hline
Greedy (BL)      & 38.8 & 75.3 & 60.3 & 55.2 & 90.0  & 38.8 & 67.3 & 50.3 & 57.7 & 93.6 \\
Beam Search (BL) & 33.4 & 77.0 & \textbf{67.7} & 54.0 & 90.6  & 31.8 & \textbf{73.8} & 51.7 & 59.0 & 93.2 \\
TTPO             & 40.4 & 75.9 & 60.8 & \textbf{56.0} & 90.4  & 38.6 & 67.2 & 50.7 & \textbf{59.4} & \textbf{96.6} \\
TTPO + Steering  & \textbf{41.4} & 76.5 & 60.7 & 54.9 & 90.0  & \textbf{40.2} & 70.4 & 51.1 & 58.0 & 94.2 \\
TTPO + Branching & 39.2 & \textbf{77.1} & 61.3 & 55.4 & \textbf{92.2}
                 & 38.6 & 69.1 & \textbf{52.1} & 57.7 & 94.2 \\
\hline
\end{tabular}

\caption{Comparing greedy, beam search and TTPO with branching and steering.}
\label{tab:ablation_4b_one_method_two_models}
\end{table*}

\section{Result}

\subsection{Experimental Setup}
\label{sec:setup}

\paragraph{Models}
All experiments are conducted on two families of instruction tuned model: Gemma 3 series \citep{team2025gemma}, Gemma-3-4b-it and Gemma-3-12b-it, Phi 4 series \citep{abdin2024phi}, Phi-4-mini-instruct and Phi-4-reasoning-plus.

\paragraph{Baselines}
We report two standard decoding baselines.
\textbf{(1) Greedy decoding} serves as our primary baseline, representing the common single pass inference setting.
Additionally include \textbf{(2) Beam search} as a stronger search-based baseline that explores multiple candidate continuations while remaining deterministic at inference time.

\paragraph{Dataset}
To evaluate deductive logical reasoning across diverse forms of structured inference, we use five benchmark datasets:
ZebraLogic \citep{lin2025zebralogic}, BIG-Bench Hard (deductive subset) \citep{suzgun2023challenging}, RuleBERT \citep{saeed2021rulebert}, LogiQA 2.0 \citep{liu2023logiqa}, and ProntoQA \citep{saparov2022language}.
These datasets collectively cover multi-step deduction, rule based inference, and formal/implicit logical transitions, providing a comprehensive testbed for evaluating robustness at logical junctions.

We additionally use OpenThought only for extracting the gradient-based steering vector (Section~\ref{sec:grad_steering}). Importantly, OpenThought \citep{guha2025openthoughts} is not used for inference time generation, candidate selection, branching evaluation, or RL training; all reported results are produced solely by the four target models listed above.

\paragraph{Method Configurations}
Detailed hyperparameters and implementation settings for \textsc{Steering}, \textsc{Branching}, and \textsc{TTPO} are provided in Appendix~\ref{sec:experimental_setup} and ~\ref{sec:prompt_template}.

\subsection{Main Results}
\label{sec:main_results}

Table~\ref{tab:main_results} summarizes results across five benchmarks, comparing deterministic decoding baselines against our connective centric methods spanning different stages:

\paragraph{Greedy is fragile, and global search is not uniformly helpful.}
Beam search improves performance on several datasets, but it can also degrade accuracy on others, notably at ZebraLogic. This reinforces the practical limitation that the best decoding strategy remains task dependent, and stronger global search does not guarantee a better reasoning trajectory under strict evaluation.
We additionally compare with self-consistency ($n=5$) in Table~\ref{tab:perf_eff_4b} (Appendix), where our methods achieve comparable average accuracy at substantially lower compute cost (Section~\ref{sec:acc_eff_tradeoff}).

\paragraph{Inference time interventions recover greedy failures at logical pivots.}
On the smaller models, our inference time methods yield consistent gains at connective junctions.
For Gemma-3-4b-it, localized branching achieves the best ZebraLogic score (38.8 vs.\ 42.0) and improves LogiQA (55.2 vs.\ 56.8), indicating that resolving ambiguity only at connective forking points can recover failures without global search.
Steering also improves over greedy across both 4B models (e.g., ZebraLogic 39.0/39.6, BBH 68.8 on Phi), suggesting that a lightweight representation shift can stabilize local transitions even under greedy decoding.

\paragraph{TTPO improves greedy behavior with minimal, localized optimization.}
TTPO consistently improves or matches greedy on most settings, while remaining strictly focused on the single token decision at logical pivots.
Notably, TTPO is particularly strong on Phi-4-mini-instruct for ProntoQA (93.6 vs.\ 96.6) and LogiQA (57.7 vs.\ 59.4), demonstrating that refining \emph{connective logits} alone can yield meaningful end task gains.
For the larger Gemma-3-12b-it model, TTPO achieves the best ZebraLogic performance (53.2 vs.\ 55.6), suggesting that token level transition refinement remains beneficial even when the base model is already strong.

\paragraph{Ablation and Method Complementarity.}
To understand how our three interventions interact, we report additional ablations on the 4B models in Table~\ref{tab:ablation_4b_one_method_two_models}. The results indicate that the proposed methods are not merely redundant variants of test time scaling, but provide complementary control signals at different stages of generation.

TTPO alone improves greedy decoding on multiple benchmarks, confirming that refining connective level decisions translates to end task gains. Combining TTPO with steering further increases robustness where greedy trajectories are brittle at local transitions, while combining TTPO with branching benefits tasks where connective ambiguity is frequent and lookahead is informative.

\section{Analysis}

\subsection{The Relationship between Connective Density and Methodological Gains}
\begin{figure}[ht]
\centering
\includegraphics[width=1.0\linewidth]{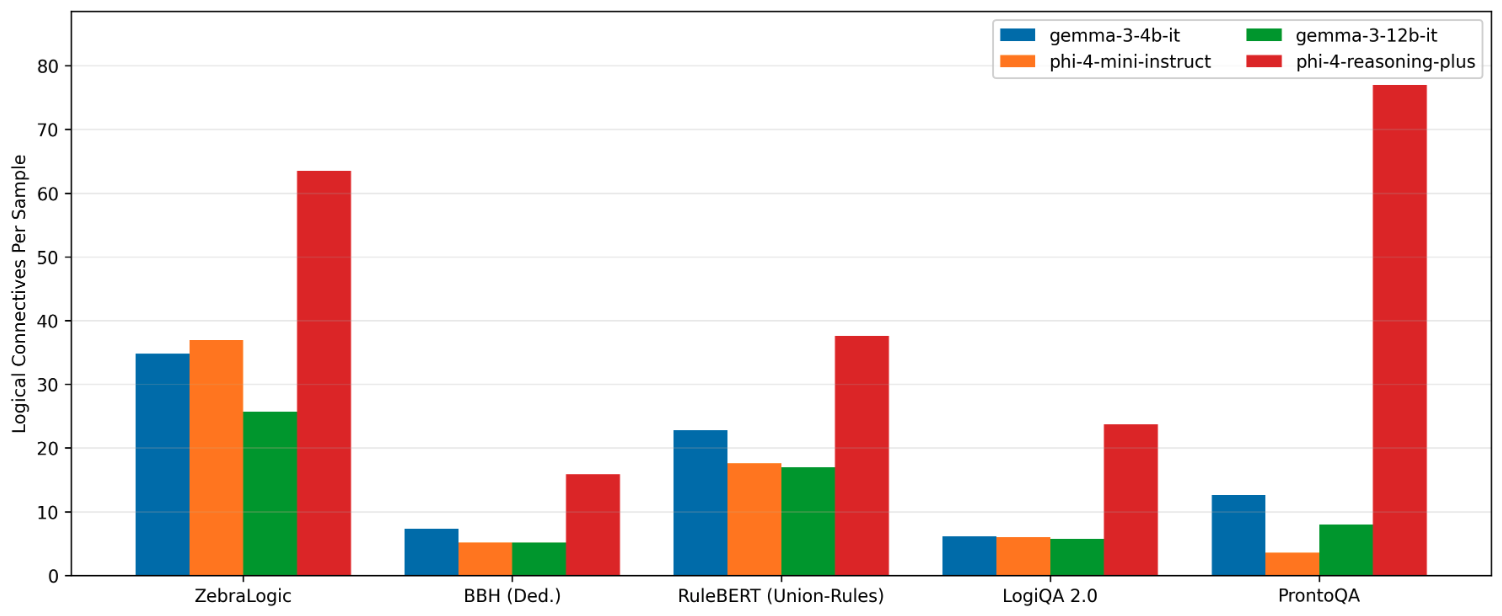}
\caption{Average count of logical connectives per sample across benchmarks under greedy decoding.}
\label{fig:logical_connectives_per_sample}
\end{figure}

Our methods trigger only at explicit connective pivots; thus, gains correlate with connective density (Figure~\ref{fig:logical_connectives_per_sample}).
Benchmarks with sparse connectives (e.g., BBH, LogiQA 2.0) offer fewer intervention opportunities, limiting improvements even when overall reasoning remains imperfect.

\subsection{Discriminative Performance of Lookahead Evaluation}
\label{sec:discriminative_eval}

\paragraph{The Sparsity of Logical Transitions} 
As shown in Figure~\ref{fig:logical_connectives_per_sample}, LogiQA 2.0 and BBH contain fewer explicit logical connectives per sample than ZebraLogic, limiting intervention opportunities for Steering and Branching. This limits the achievable gains, since failures outside explicit connective transitions are less likely to be corrected by our interventions.

To validate the reliability of our branching metric, we evaluated whether the lookahead score effectively identifies the ground truth logical connective. We ranked a candidate set containing the ground-truth token alongside the model's top-$K$ predictions based on our proposed metric.

\begin{table}[h]
\centering
\small
\begin{tabular}{lc}
\hline
\textbf{Model} & \textbf{Match Rate (\%)} \\
\hline
Gemma-3-4b-it & 73.41 \\
Phi-4-mini-instruct & 69.14 \\
\hline
\end{tabular}
\caption{Match rate of the lookahead selection mechanism against ground-truth logical connectives.}
\label{tab:discriminative_acc}
\end{table}

As shown in Table~\ref{tab:discriminative_acc}, our metric demonstrates strong discriminative capability, aligning with the ground truth in 73.4\% and 69.1\% of cases for Gemma and Phi models, respectively. This confirms that the lookahead score acts as a robust proxy for logical validity, successfully guiding the model toward the correct reasoning path.

\subsection{Distributional Sharpening at Logical Pivots}

\begin{figure}[ht]
\centering
\includegraphics[width=1.0\linewidth]{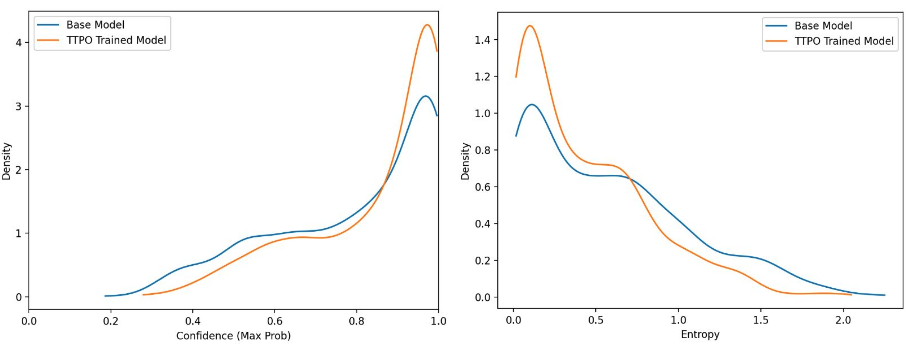}
\caption{Distribution of prediction confidence \textit{(left)} and entropy \textit{(right)} at logical connective positions. TTPO trained models exhibit a significant shift toward higher certainty and lower ambiguity.}
\label{fig:ttpo_distribution}
\end{figure}

To investigate how TTPO refines the model's internal decision making, we analyze the probability distribution at the specific positions where logical connectives ($\mathcal{S}_l$) are predicted. Figure~\ref{fig:ttpo_distribution} illustrates the density of prediction confidence and entropy for Gemma-3-4b-it.

The results demonstrate that TTPO effectively sharpens the probability distribution at logical pivots. While the baseline model exhibits high entropy and low confidence at these forking points, TTPO trained models show a marked shift toward higher certainty and lower entropy. This distributional sharpening indicates that our surgical optimization successfully resolves the structural ambiguity inherent in logical transitions. By increasing the margin between the optimal connective and sub-optimal alternatives, the model commits decisively to a single reasoning path, thereby mitigating error propagation during greedy decoding without the need for exhaustive search.

\subsection{Accuracy and Efficiency Trade-off}
\label{sec:acc_eff_tradeoff}

Table~\ref{tab:eff_cost_4b_combined} shows that global test time scaling baselines incur $3{\sim}6\times$ higher token cost and up to $2.8{\sim}4.1\times$ higher latency than greedy on 4B models, whereas our connective centric methods remain close to single pass decoding.
As shown in Table~\ref{tab:perf_eff_4b}, on Gemma-3-4b-it our best method per task achieves higher average accuracy than self-consistency ($n{=}5$) (65.7 vs.\ 65.2) at $\le$1.45$\times$ greedy cost, compared to 2.90$\times$ for self-consistency. On Phi-4-mini-instruct, self-consistency achieves marginally higher accuracy (63.8 vs.\ 63.3) but at 4.37$\times$ token cost---a substantial overhead for a 0.5 point gain. Overall, targeting connective pivots approximates the benefits of global sampling at a fraction of the compute.

\section{Conclusion}

We identify logical connectives as critical pivots that dictate reasoning trajectories. Our diagnostic analysis reveals that these tokens concentrate model uncertainty and exhibit disproportionate causal leverage over reasoning outcomes. Rather than relying on global optimization, we demonstrate that intervening only at these transitions, via activation-level steering, localized branching, and targeted preference optimization, is sufficient to enhance reasoning performance while preserving near-greedy efficiency, suggesting that linguistically grounded token-level intervention offers a complementary paradigm to global test time scaling.

\section*{Limitations}

Our approach faces limitations stemming from its reliance on explicit linguistic markers. First, by utilizing a predefined taxonomy of logical connectives, our method cannot address implicit reasoning steps where transitions occur without specific discourse markers. Second, the detection mechanism is sensitive to tokenizer artifacts and polysemy, which may introduce alignment errors across different models. Finally, the effectiveness of our framework is bounded by the density of connectives; tasks with sparse explicit transitions, such as BBH, show limited performance gains compared to linguistically rich datasets like ZebraLogic.


\bibliography{custom}

@article{turner2308steering,
  title={Steering language models with activation engineering, 2023},
  author={Turner, Alexander Matt and Thiergart, Lisa and Leech, Gavin and Udell, David and Vazquez, Juan J and Mini, Ulisse and MacDiarmid, Monte},
  year={2023},
}

@article{creswell2022selection,
  title={Selection-inference: Exploiting large language models for interpretable logical reasoning},
  author={Creswell, Antonia and Shanahan, Murray and Higgins, Irina},
  journal={arXiv preprint arXiv:2205.09712},
  year={2022}
}

@article{creswell2022faithful,
  title={Faithful reasoning using large language models},
  author={Creswell, Antonia and Shanahan, Murray},
  journal={arXiv preprint arXiv:2208.14271},
  year={2022}
}

@article{bogdan2025thought,
  title={Thought Anchors: Which LLM Reasoning Steps Matter?},
  author={Bogdan, Paul C and Macar, Uzay and Nanda, Neel and Conmy, Arthur},
  journal={arXiv preprint arXiv:2506.19143},
  year={2025}
}

@article{chen2023accelerating,
  title={Accelerating large language model decoding with speculative sampling},
  author={Chen, Charlie and Borgeaud, Sebastian and Irving, Geoffrey and Lespiau, Jean-Baptiste and Sifre, Laurent and Jumper, John},
  journal={arXiv preprint arXiv:2302.01318},
  year={2023}
}

@article{chen2025towards,
  title={Towards reasoning era: A survey of long chain-of-thought for reasoning large language models},
  author={Chen, Qiguang and Qin, Libo and Liu, Jinhao and Peng, Dengyun and Guan, Jiannan and Wang, Peng and Hu, Mengkang and Zhou, Yuhang and Gao, Te and Che, Wanxiang},
  journal={arXiv preprint arXiv:2503.09567},
  year={2025}
}

@article{fu2025deep,
  title={Deep think with confidence},
  author={Fu, Yichao and Wang, Xuewei and Tian, Yuandong and Zhao, Jiawei},
  journal={arXiv preprint arXiv:2508.15260},
  year={2025}
}

@article{team2025gemma,
  title={Gemma 3 technical report},
  author={Team, Gemma and Kamath, Aishwarya and Ferret, Johan and Pathak, Shreya and Vieillard, Nino and Merhej, Ramona and Perrin, Sarah and Matejovicova, Tatiana and Ram{\'e}, Alexandre and Rivi{\`e}re, Morgane and others},
  journal={arXiv preprint arXiv:2503.19786},
  year={2025}
}

@article{guha2025openthoughts,
  title={OpenThoughts: Data Recipes for Reasoning Models},
  author={Guha, Etash and Marten, Ryan and Keh, Sedrick and Raoof, Negin and Smyrnis, Georgios and Bansal, Hritik and Nezhurina, Marianna and Mercat, Jean and Vu, Trung and Sprague, Zayne and others},
  journal={arXiv preprint arXiv:2506.04178},
  year={2025}
}

@misc{he2025codeiocondensingreasoning,
      title={CodeI/O: Condensing Reasoning Patterns via Code Input-Output Prediction}, 
      author={Junxian He and Yu Wu and Junlong Li and Daya Guo and Dejian Yang and Runxin Xu},
      year={2025},
      archivePrefix={arXiv},
      primaryClass={cs.CL},
}

@inproceedings{lightman2023let,
  title={Let's verify step by step},
  author={Lightman, Hunter and Kosaraju, Vineet and Burda, Yuri and Edwards, Harrison and Baker, Bowen and Lee, Teddy and Leike, Jan and Schulman, John and Sutskever, Ilya and Cobbe, Karl},
  booktitle={The Twelfth International Conference on Learning Representations},
  year={2023}
}

@article{lei2024boosting,
  title={Boosting logical fallacy reasoning in LLMs via logical structure tree},
  author={Lei, Yuanyuan and Huang, Ruihong},
  journal={arXiv preprint arXiv:2410.12048},
  year={2024}
}

@article{lin2025zebralogic,
  title={Zebralogic: On the scaling limits of llms for logical reasoning},
  author={Lin, Bill Yuchen and Bras, Ronan Le and Richardson, Kyle and Sabharwal, Ashish and Poovendran, Radha and Clark, Peter and Choi, Yejin},
  journal={arXiv preprint arXiv:2502.01100},
  year={2025}
}

@article{lin2024critical,
  title={Critical Tokens Matter: Token-Level Contrastive Estimation Enhances LLM's Reasoning Capability},
  author={Lin, Zicheng and Liang, Tian and Xu, Jiahao and Lin, Qiuzhi and Wang, Xing and Luo, Ruilin and Shi, Chufan and Li, Siheng and Yang, Yujiu and Tu, Zhaopeng},
  journal={arXiv preprint arXiv:2411.19943},
  year={2024}
}

@article{liu2025safe,
  title={Safe: Enhancing Mathematical Reasoning in Large Language Models via Retrospective Step-aware Formal Verification},
  author={Liu, Chengwu and Yuan, Ye and Yin, Yichun and Xu, Yan and Xu, Xin and Chen, Zaoyu and Wang, Yasheng and Shang, Lifeng and Liu, Qun and Zhang, Ming},
  journal={arXiv preprint arXiv:2506.04592},
  year={2025}
}

@article{liu2023logiqa,
  title={Logiqa 2.0—an improved dataset for logical reasoning in natural language understanding},
  author={Liu, Hanmeng and Liu, Jian and Cui, Leyang and Teng, Zhiyang and Duan, Nan and Zhou, Ming and Zhang, Yue},
  journal={IEEE/ACM Transactions on Audio, Speech, and Language Processing},
  volume={31},
  pages={2947--2962},
  year={2023},
  publisher={IEEE}
}

@misc{luo2025codethinkthink,
      title={Code to Think, Think to Code: A Survey on Code-Enhanced Reasoning and  Reasoning-Driven Code Intelligence in LLMs}, 
      author={Jiebo Luo and Xin Qian and Julian McAuley and Yuwei Cao and Daoan Zhang and Tianyang Liu and Xiaoyi Liu and Antoine Simoulin and Grey Yang and Dayu Yang and Zhaopu Teng},
      year={2025},
      archivePrefix={arXiv},
      primaryClass={cs.CL},
}

@article{abdin2024phi,
  title={Phi-4 technical report},
  author={Abdin, Marah and Aneja, Jyoti and Behl, Harkirat and Bubeck, S{\'e}bastien and Eldan, Ronen and Gunasekar, Suriya and Harrison, Michael and Hewett, Russell J and Javaheripi, Mojan and Kauffmann, Piero and others},
  journal={arXiv preprint arXiv:2412.08905},
  year={2024}
}

@article{mcginness2024automated,
  title={Automated theorem provers help improve large language model reasoning},
  author={McGinness, Lachlan and Baumgartner, Peter},
  journal={arXiv preprint arXiv:2408.03492},
  year={2024}
}

@inproceedings{suzgun2023challenging,
  title={Challenging big-bench tasks and whether chain-of-thought can solve them},
  author={Suzgun, Mirac and Scales, Nathan and Sch{\"a}rli, Nathanael and Gehrmann, Sebastian and Tay, Yi and Chung, Hyung Won and Chowdhery, Aakanksha and Le, Quoc and Chi, Ed and Zhou, Denny and others},
  booktitle={Findings of the Association for Computational Linguistics: ACL 2023},
  pages={13003--13051},
  year={2023}
}

@article{saeed2021rulebert,
  title={RuleBERT: Teaching soft rules to pre-trained language models},
  author={Saeed, Mohammed and Ahmadi, Naser and Nakov, Preslav and Papotti, Paolo},
  journal={arXiv preprint arXiv:2109.13006},
  year={2021}
}

@inproceedings{rimsky2024steering,
  title={Steering llama 2 via contrastive activation addition},
  author={Rimsky, Nina and Gabrieli, Nick and Schulz, Julian and Tong, Meg and Hubinger, Evan and Turner, Alexander},
  booktitle={Proceedings of the 62nd Annual Meeting of the Association for Computational Linguistics (Volume 1: Long Papers)},
  pages={15504--15522},
  year={2024}
}

@inproceedings{pan2023logic,
  title={Logic-lm: Empowering large language models with symbolic solvers for faithful logical reasoning},
  author={Pan, Liangming and Albalak, Alon and Wang, Xinyi and Wang, William},
  booktitle={Findings of the Association for Computational Linguistics: EMNLP 2023},
  pages={3806--3824},
  year={2023}
}

@article{panickssery2312steering,
  title={Steering llama 2 via contrastive activation addition, 2024},
  author={Panickssery, Nina and Gabrieli, Nick and Schulz, Julian and Tong, Meg and Hubinger, Evan and Turner, Alexander Matt},
  year={2024},
  volume={3},
}

@inproceedings{robaldo2008penn,
  title={The penn discourse treebank 2.0},
  author={Robaldo, Livio},
  booktitle={Lrec},
  year={2008}
}

@article{saparov2022language,
  title={Language models are greedy reasoners: A systematic formal analysis of chain-of-thought},
  author={Saparov, Abulhair and He, He},
  journal={arXiv preprint arXiv:2210.01240},
  year={2022}
}

@article{wang2025beyond,
  title={Beyond the 80/20 rule: High-entropy minority tokens drive effective reinforcement learning for llm reasoning},
  author={Wang, Shenzhi and Yu, Le and Gao, Chang and Zheng, Chujie and Liu, Shixuan and Lu, Rui and Dang, Kai and Chen, Xionghui and Yang, Jianxin and Zhang, Zhenru and others},
  journal={arXiv preprint arXiv:2506.01939},
  year={2025}
}

@article{wang2022self,
  title={Self-consistency improves chain of thought reasoning in language models},
  author={Wang, Xuezhi and Wei, Jason and Schuurmans, Dale and Le, Quoc and Chi, Ed and Narang, Sharan and Chowdhery, Aakanksha and Zhou, Denny},
  journal={arXiv preprint arXiv:2203.11171},
  year={2022}
}

@misc{wei2024llmmastermindsurvey,
      title={LLM as a Mastermind: A Survey of Strategic Reasoning with Large Language
  Models}, 
      author={Furu Wei and Wenshan Wu and Shaoguang Mao and Yadong Zhang and Yan Xia and Tao Ge and Man Lan and Xun Wang and Ting Song and Adrian de Wynter},
      year={2024},
      archivePrefix={arXiv},
      primaryClass={cs.CL},
}

@article{wei2022chain,
  title={Chain-of-thought prompting elicits reasoning in large language models},
  author={Wei, Jason and Wang, Xuezhi and Schuurmans, Dale and Bosma, Maarten and Xia, Fei and Chi, Ed and Le, Quoc V and Zhou, Denny and others},
  journal={Advances in neural information processing systems},
  volume={35},
  pages={24824--24837},
  year={2022}
}

@Article{Wei2022ChainOT,
 author = {Jason Wei and Xuezhi Wang and Dale Schuurmans and Maarten Bosma and Ed H. Chi and F. Xia and Quoc Le and Denny Zhou},
 booktitle = {Neural Information Processing Systems},
 journal = {ArXiv},
 title = {Chain of Thought Prompting Elicits Reasoning in Large Language Models},
 volume = {abs/2201.11903},
 year = {2022}
}

@article{yang2023logical,
  title={Logical reasoning over natural language as knowledge representation: A survey},
  author={Yang, Zonglin and Du, Xinya and Mao, Rui and Ni, Jinjie and Cambria, Erik},
  journal={arXiv preprint arXiv:2303.12023},
  year={2023}
}

@article{yao2023tree,
  title={Tree of thoughts: Deliberate problem solving with large language models, 2023},
  author={Yao, Shunyu and Yu, Dian and Zhao, Jeffrey and Shafran, Izhak and Griffiths, Thomas L and Cao, Yuan and Narasimhan, Karthik},
  volume={3},
  pages={1},
  year={2023}
}

@misc{zhang2025systemsystemsurvey,
      title={From System 1 to System 2: A Survey of Reasoning Large Language Models}, 
      author={Jiaxin Zhang and Zhijiang Guo and Haotian Xu and Yingying Zhang and Jiahua Dong and Yuxuan Yao and Le Song and Cheng-Lin Liu and Junhao Zheng and Fei Yin and Duzhen Zhang and Xiuyi Chen and Zhong-Zhi Li and Ming-Liang Zhang and Zengyan Liu and Pei-Jie Wang},
      year={2025},
      archivePrefix={arXiv},
      primaryClass={cs.AI},
}

@misc{zhang2025empoweringllmslogical,
      title={Empowering LLMs with Logical Reasoning: A Comprehensive Survey}, 
      author={Kun Zhang and Zhouchen Lin and Haoxuan Li and Robert van Rooij and Fengxiang Cheng and Fenrong Liu},
      year={2025},
      archivePrefix={arXiv},
      primaryClass={cs.AI},
}

@misc{zhang2025logicalreasoninglarge,
      title={Logical Reasoning in Large Language Models: A Survey}, 
      author={Yue Zhang and Chaoli Zhang and Hanmeng Liu and Ruoxi Ning and Mengru Ding and Zhizhang Fu and Xiaozhang Liu},
      year={2025},
      archivePrefix={arXiv},
      primaryClass={cs.AI},
}

@inproceedings{zhao2024lookahead,
  title={Lookahead: An inference acceleration framework for large language model with lossless generation accuracy},
  author={Zhao, Yao and Xie, Zhitian and Liang, Chen and Zhuang, Chenyi and Gu, Jinjie},
  booktitle={Proceedings of the 30th ACM SIGKDD Conference on Knowledge Discovery and Data Mining},
  pages={6344--6355},
  year={2024}
}

@article{zur2025language,
  title={Are language models aware of the road not taken? Token-level uncertainty and hidden state dynamics},
  author={Zur, Amir and Geiger, Atticus and Lubana, Ekdeep Singh and Bigelow, Eric},
  journal={arXiv preprint arXiv:2511.04527},
  year={2025}
}

\appendix

\section{Logical Connective Set}
\label{sec:logical_connectives}

To systematically identify the pivotal points in reasoning chains, we define a set of target logical connectives, denoted as $\mathcal{S}_{l}$. Our taxonomy is grounded in the discourse relation framework proposed by \citet{robaldo2008penn} and further refined by recent work in logical fallacy detection \citep{lei2024boosting}.

We categorize logical relations into ten distinct types that are essential for multi step deduction: \textit{Conjunction, Alternative, Restatement, Instantiation, Contrast, Concession, Analogy, Temporal, Condition,} and \textit{Causal} relations. The set $\mathcal{S}_{l}$ is constructed by aggregating explicit discourse connectives associated with these categories. In this process, we explicitly exclude high frequency ambiguous tokens such as ``and'' and ``or''. While they theoretically denote logical relations, they predominantly function as syntactic coordinators in natural language. Pruning these tokens ensures that our framework targets explicit deductive transitions rather than simple grammatical conjunctions.

\paragraph{Multi-token connectives.}
Importantly, many discourse connectives are realized as multitoken phrases (e.g., ``as a result'', ``in contrast'', ``on the other hand'') rather than a single token under subword tokenization. Therefore, our intervention target $\mathcal{S}_{l}$ is defined at the string/phrase level, and we implement connective detection using suffix matching over the most recent $n$ generated tokens. Concretely, at each decoding step we detokenize the last $n$ tokens (we use $n$ large enough to cover the longest connective phrase in $\mathcal{S}_{l}$) and check whether any connective phrase in $\mathcal{S}_{l}$ matches the resulting suffix. This allows \textsc{Steering} and \textsc{Branching} to reliably trigger at both single token and multi token connective junctions.

Table~\ref{tab:connective_list} presents the categorization and representative examples of the logical connectives used in our experiments. These connectives serve as the trigger points for our \textsc{Steering}, \textsc{Branching}, and \textsc{TTPO} mechanisms.

\section{Entropy Analysis}
\label{sec:entropy_analysis}

To ensure that our findings regarding the high entropy of logical connectives are robust and not artifacts of a specific threshold ($\tau$), we present detailed supplementary analyses conducted on the ZebraLogic dataset.

\subsection{Threshold Free Quantile Enrichment}
We first conducted a threshold free quantile enrichment analysis. Let $H_q$ be the $q$-th quantile of the entropy distribution over all generated tokens. We measure the proportion of logical connectives in the overall generation (Base \%) and compare it to their proportion within the high-entropy tail (Tail \%). 

As shown in Table~\ref{tab:quantile_enrichment}, even though connectives constitute only a small fraction of all tokens (4.05\% for Gemma, 6.73\% for Phi), they are strongly over-represented in every high-entropy tail. For instance, in the top 1\% most uncertain tokens ($q=0.99$), connectives are enriched by 5.77$\times$ and 2.01$\times$ for Gemma and Phi, respectively. This confirms that connectives intrinsically dominate the model's structural uncertainty, independent of any specific threshold.

\begin{table*}[h]
\centering
\small
\setlength{\tabcolsep}{4pt}
\begin{tabular}{lcccc}
\hline
\textbf{} & \textbf{Tail Quantile ($q$)} & \textbf{Base Conn. (\%)} & \textbf{Tail Conn. (\%)} & \textbf{Enrichment} \\
\hline
Gemma-3-4b-it & 0.90 & 4.05 & 18.42 & 4.55$\times$ \\
Gemma-3-4b-it & 0.95 & 4.05 & 22.85 & 5.64$\times$ \\
Gemma-3-4b-it & 0.99 & 4.05 & 23.37 & 5.77$\times$ \\
\hline
Phi-4-mini-instruct & 0.90 & 6.73 & 18.58 & 2.76$\times$ \\
Phi-4-mini-instruct & 0.95 & 6.73 & 18.17 & 2.70$\times$ \\
Phi-4-mini-instruct & 0.99 & 6.73 & 13.54 & 2.01$\times$ \\
\hline
\end{tabular}
\caption{Threshold free quantile enrichment of logical connectives in the high-entropy tails.}
\label{tab:quantile_enrichment}
\end{table*}

\subsection{Entropy Across Token Categories}
To verify that this uncertainty is unique to connectives and not a generic property of all functional tokens, we computed the High Entropy Rate ($R_{HE}$) for various token categories across a broad sweep of $\tau$ thresholds. 

As shown in Table~\ref{tab:tau_sweep}, logical connectives consistently maintain the highest $R_{HE}$ among all categories across all thresholds. While quantifiers in Phi-4-mini show a relatively high $R_{HE}$ at certain thresholds, they constitute an extremely rare portion of the output ($\sim$0.36\% of all tokens) and do not function as inter-step relation operators that direct logical trajectories. These results empirically validate our focus on logical connectives as the primary and most actionable intervention targets for reasoning.

\begin{table*}[h]
\centering
\small
\setlength{\tabcolsep}{4pt}
\begin{tabular}{lcccccc}
\hline
\multicolumn{7}{}{\textbf{Gemma-3-4b-it}} \\
\hline
\textbf{Threshold ($\tau$)} & \textbf{Conn.} & \textbf{Non-conn.} & \textbf{Negation} & \textbf{Quantifier} & \textbf{Number} & \textbf{Punct.} \\
\hline
0.5 & 64.5\% & 17.7\% & 25.1\% & 13.7\% & 5.8\% & 15.5\% \\
1.0 (Default) & 40.9\% & 7.1\% & 12.3\% & 6.9\% & 1.0\% & 3.7\% \\
1.5 & 21.0\% & 2.7\% & 3.5\% & 1.7\% & 0.1\% & 0.6\% \\
2.0 & 6.6\% & 0.9\% & 0.9\% & 0.4\% & 0.0\% & 0.1\% \\
\hline
\hline
\multicolumn{7}{}{\textbf{Phi-4-mini-instruct}} \\
\hline
\textbf{Threshold ($\tau$)} & \textbf{Conn.} & \textbf{Non-conn.} & \textbf{Negation} & \textbf{Quantifier} & \textbf{Number} & \textbf{Punct.} \\
\hline
0.5 & 70.1\% & 33.9\% & 41.1\% & 53.5\% & 32.8\% & 40.8\% \\
1.0 (Default) & 56.8\% & 23.0\% & 34.5\% & 42.8\% & 20.6\% & 25.3\% \\
1.5 & 43.0\% & 14.6\% & 26.1\% & 30.3\% & 11.3\% & 12.0\% \\
2.0 & 27.9\% & 8.8\% & 16.3\% & 22.3\% & 4.3\% & 5.1\% \\
\hline
\end{tabular}
\caption{High Entropy Rate ($R_{HE}$) comparison across different token categories and $\tau$ thresholds. Logical connectives (Conn.) consistently show the highest structural uncertainty.}
\label{tab:tau_sweep}
\end{table*}

\begin{table*}[h]
\centering
\small
\renewcommand{\arraystretch}{1.2}
\begin{tabular}{p{0.3\linewidth} p{0.6\linewidth}}
\hline
Logical Relations & Relation Connectives \\
\hline
Conjunction & as well as, as well, also, separately \\
\hline
Alternative & either, instead, alternatively, else, neither \\
\hline
Restatement & specifically, particularly, in particular, besides, additionally, in addition, moreover, furthermore,
plus, not only, indeed, in other words, in fact, in short, in the end, overall, in summary, in details \\
\hline
Instantiation & for example, for instance, such as, including, as an example, an as instance, for one thing \\
\hline
Contrast & but, however, yet, while, unlike, rather, rather than, in comparison, by comparison, on the other hand,
on the contrary, contrary to, in contrast, by contrast, whereas, conversely \\
\hline
Concession & although, though, despite, despite of, in spite of, regardless, regardless of, nevertheless,
nonetheless, even if, even though, even as, even when, even after, even so, no matter \\
\hline
Analogy & likewise, similarly, as if, as though, just as, just like, namely \\
\hline
Temporal & during, before, after, when, as soon as, then, next, until, till, meanwhile, in turn, meantime, afterwards,
simultaneously, at the same time, beforehand, previously, earlier, later, thereafter, finally, ultimately \\
\hline
Condition & if, as long as, unless, otherwise, except, whenever, whichever, once, only if, only when, depend on \\
\hline
Causal & because, cause, as a result, result in, due to, therefore, hence, thus, thereby, since, now that,
consequently, in consequence, in order to, so as to, so that, why, for, accordingly, given, turn out \\
\hline
\end{tabular}
\caption{Taxonomy of logical relations and examples of logical connectives included in $\mathcal{S}_{l}$. This set acts as the intervention target.}
\label{tab:connective_list}
\end{table*}

\section{Experimental Setup}
\label{sec:experimental_setup}

\subsection{Logical Steering}
For gradient based steering, we utilized the \textbf{OpenThought} dataset to extract reasoning vectors. We computed the steering vector from the activations of the \textbf{penultimate layer} (the second to last layer) of the model. This layer was chosen to capture high level semantic reasoning features before they are projected into the vocabulary space. During inference, we applied the steering vector with an injection coefficient of $\alpha = 0.5$ across all models.

\subsection{Logical Branching}
The branching mechanism was configured with a lookahead depth of $L=20$. 
To ensure computational efficiency, we trigger branching only at steps where the greedy top 1 token is itself a logical connective, i.e., $w_{\text{top1}} \in \mathcal{S}_{l}$. 
Even then, we apply an additional ambiguity filter: branching is activated only when at least two candidates from our predefined connective set $\mathcal{S}_{l}$ appear within the top $20$ token predictions. 
This two-stage criterion ensures that branching is performed only at genuine decision points, connective pivots where the model exhibits structural uncertainty, while avoiding unnecessary lookahead on non-connective or low ambiguity steps.

\subsection{Targeted Transition Preference Optimization (TTPO)}

\subsubsection{Data Construction}
\label{sec:data_construction}
To train the TTPO objective, we constructed a pairwise preference dataset focused on logical transitions. We first split the \textbf{ZebraLogic} dataset into training and test sets with an 8:2 ratio. Using the training split, we performed inference to identify logical pivot points.

\begin{algorithm}[t]
\caption{TTPO Dataset Construction}
\label{alg:ttpo_data}
\small
\begin{algorithmic}[1]
\Require Training set $\mathcal{D}_{\text{train}}$, connective pool $\mathcal{S}_{l}$, model $\mathcal{M}$
\Ensure Preference pairs $\mathcal{D}_{\text{pair}}$
\State $\mathcal{D}_{\text{pair}} \gets \emptyset$
\ForAll{$(x, y_{\text{gold}}) \in \mathcal{D}_{\text{train}}$}
    \State $t \gets \textsc{FindPivot}(\mathcal{M}, x)$
    \State $w_{\text{greedy}} \gets \textsc{GreedyToken}(\mathcal{M}, x_{\le t})$
    \State $C \gets \{w_{\text{greedy}}\} \cup \textsc{Sample}(\mathcal{S}_{l}, 5)$
    \State $\mathcal{P}_{\text{pos}} \gets \emptyset;\;\; \mathcal{P}_{\text{neg}} \gets \emptyset$
    \ForAll{$c \in C$}
        \State $y_{\text{gen}} \gets \textsc{GreedyDecode}(\mathcal{M}, x_{\le t} \Vert c)$
        \If{$\textsc{CheckAnswer}(y_{\text{gen}}, y_{\text{gold}})$}
            \State $\mathcal{P}_{\text{pos}} \gets \mathcal{P}_{\text{pos}} \cup \{c\}$
        \Else
            \State $\mathcal{P}_{\text{neg}} \gets \mathcal{P}_{\text{neg}} \cup \{c\}$
        \EndIf
    \EndFor
    \If{$\mathcal{P}_{\text{pos}} \neq \emptyset \;\And\; \mathcal{P}_{\text{neg}} \neq \emptyset$}
        \State $w_c \gets \textsc{Select}(\mathcal{P}_{\text{pos}})$;\;\;
               $w_r \gets \textsc{Select}(\mathcal{P}_{\text{neg}})$
        \State $\mathcal{D}_{\text{pair}} \gets \mathcal{D}_{\text{pair}} \cup \{(x, w_c, w_r)\}$
    \EndIf
\EndFor
\end{algorithmic}
\end{algorithm}

The data construction process is detailed in Algorithm~\ref{alg:ttpo_data}. When the model encounters a logical connective during generation, we create branches using the original greedy token and 5 randomly sampled connectives from $\mathcal{S}_{l}$. For each branch, we continue generation using greedy decoding until the EOS token. We then verify the correctness of each generated path. If we identify a pair where one transition leads to the correct answer ($w_c$) and another leads to an incorrect answer ($w_r$), we form a preference pair $(x, w_c, w_r)$. 
This dataset is also utilized for the discriminative analysis in Section~\ref{sec:discriminative_eval}.

\subsubsection{Training Configuration}
We fine tuned the models using the constructed preference pairs for 3 epochs with a batch size of 1. To prevent catastrophic forgetting and ensure stable convergence on the specific transition tokens, we used a low learning rate: $1\times 10^{-6}$ for the 4B parameters models and $1\times 10^{-7}$ for the 13B models.

\subsection{Benchmark}
\label{sec:benchmark}

\begin{table}[t]
\centering
\small
\setlength{\tabcolsep}{6pt}
\begin{tabular}{p{0.3\linewidth}rrp{0.22\linewidth}}
\hline
Dataset & Train & Test & Task \\
\hline
ZebraLogic & 2700 & 500 & Multi Class \\
BIG Bench Hard & -- & 1000 & Multi Class \\
RuleBERT-Union & -- & 1000 & Binary Class \\
LogiQA 2.0 & -- & 1000 & Binary Class \\
ProntoQA & -- & 500 & Binary Class \\
\hline
\end{tabular}
\caption{Benchmark datasets and evaluation splits used}
\label{tab:benchmark_stats}
\end{table}

We summarize benchmark statistics in Table~\ref{tab:benchmark_stats}. 
For ZebraLogic, the training split is used exclusively for TTPO preference pair data construction (Section~\ref{sec:token_optimization}); it is not used for reporting evaluation results.

\section{Prompt Template}
\label{sec:prompt_template}

\begin{center}
\subsection{Rulebert-Union}
\label{fig:rulebert_union_prompt}
\begin{promptlisting}[Rulebert-Union Prompt]
|\textcolor{blue}{[System Instruction]}|
You are an expert in logical reasoning and reading comprehension. Your task solve questions.

Follow these steps strictly:
1. Reasoning step by step.
2. Select the answer in the format '/boxed{ANSWER}'. for example, if the answer is option A, the output should be '/boxed{A}'

|\textcolor{blue}{[User Prompt]}|
# Context:
|\textcolor{red}{[context]}|

# Question:
|\textcolor{red}{[question]}|

# Options:
A. True
B. False

Think step by step.

\end{promptlisting}
\captionof{figure}{Rulebert-Union prompt template}
\end{center}

\begin{center}
\subsection{Logi QA 2.0}
\begin{promptlisting}[Logi QA 2.0 Prompt]
|\textcolor{blue}{[System Instruction]}|
You are an expert in logical reasoning and reading comprehension. Your task solve questions.

Follow these steps strictly:
1. Reasoning step by step.
2. Select the answer in the format '/boxed{ANSWER}'. for example, if the answer is option A, the output should be '/boxed{A}'

|\textcolor{blue}{[User Prompt]}|
# Hypothesis:
|\textcolor{red}{[hypothesis]}|

# Question:
|\textcolor{red}{[question]}|

# Options:
A. not-entailment
B. entailment

Think step by step.

\end{promptlisting}
\captionof{figure}{Logi QA 2.0 prompt template}
\label{fig:logi_qa_2_0_prompt}
\end{center}

\begin{center}
\subsection{ProntoQA}
\begin{promptlisting}[ProntoQA Prompt]
|\textcolor{blue}{[System Instruction]}|
You are an expert in logical reasoning and reading comprehension. Your task solve questions.

Follow these steps strictly:
1. Reasoning step by step.
2. Output the answer in the format '/boxed{}' ANSWER is one of /boxed{A}, /boxed{B}

|\textcolor{blue}{[User Prompt]}|
# Context:
|\textcolor{red}{[context]}|

# Question:
|\textcolor{red}{[question]}|

# Options:
A. True
B. False

Think step by step.

\end{promptlisting}
\captionof{figure}{ProntoQA prompt template}
\label{fig:pronto_qa_prompt}
\end{center}

\begin{center}
\subsection{ZebraLogic}
\begin{promptlisting}[ZebraLogic Prompt]
|\textcolor{blue}{[System Instruction]}|
You are an expert at solving puzzle problems. Follow these rules strictly:
1. Solve and think step by step.
2. Do not explain anything else.
3. Give the final answer only inside /boxed{}.
- Example :
# Puzzle
...

# Question
...

# Choices
[
"Eric",
"Bob",
"Alice",
"Peter",
"Carol",
"Arnold"
]

- Answer example
# Reasoning
...

# Answer
/boxed{Bob}

|\textcolor{blue}{[User Prompt]}|
# Puzzle
|\textcolor{red}{[puzzle]}|

# Question
|\textcolor{red}{[question]}|

# Choices
|\textcolor{red}{[choices]}|

Think step by step.

\end{promptlisting}
\captionof{figure}{ZebraLogic prompt template}
\label{fig:zebralogic_prompt}
\end{center}

\begin{table*}[t]
\centering
\small
\setlength{\tabcolsep}{6pt}
\renewcommand{\arraystretch}{1.15}
\begin{tabular}{lccccc}
\hline
\multicolumn{6}{c}{Gemma-3-4b-it} \\
\hline
Dataset & Greedy & Beam & Self-Cons. (n=5) & Ours (best) & Ours method \\
\hline
ZebraLogic   & 38.8 & 33.4 & 39.6 & \textbf{42.0} & Branching \\
BBH (Ded.)   & 75.3 & \textbf{77.0} & 75.5 & 75.9 & TTPO \\
RuleBERT     & 60.3 & \textbf{67.7} & 63.9 & 63.4 & Branching \\
LogiQA 2.0   & 55.2 & 54.0 & 55.3 & \textbf{56.8} & Branching \\
ProntoQA     & 90.0 & 90.6 & \textbf{91.8} & 90.4 & TTPO \\
\hline
Avg.         & 63.9 & 64.5 & 65.2 & \textbf{65.7} & -- \\
\hline\hline
\multicolumn{6}{c}{Phi-4-mini-instruct (4B)} \\
\hline
Dataset & Greedy & Beam & Self-Cons. (n=5) & Ours (best) & Ours method \\
\hline
ZebraLogic   & 38.8 & 31.8 & \textbf{41.0} & 39.6 & Steering \\
BBH (Ded.)   & 67.3 & \textbf{73.8} & 73.6 & 69.1 & Branching \\
RuleBERT     & 50.3 & 51.7 & 48.8 & \textbf{51.9} & Branching \\
LogiQA 2.0   & 57.7 & 59.0 & 58.8 & \textbf{59.4} & TTPO \\
ProntoQA     & 93.6 & 93.2 & \textbf{96.8} & 96.6 & TTPO \\
\hline
Avg.         & 61.5 & 61.9 & \textbf{63.8} & 63.3 & -- \\
\hline
\end{tabular}
\caption{Accuracy comparison on 4B models. Self-consistency uses n=5 samples. Beam uses the same beam size as in Table~\ref{tab:main_results}.}
\label{tab:perf_eff_4b}
\end{table*}

\begin{table*}[t]
\centering
\small
\setlength{\tabcolsep}{7pt}
\renewcommand{\arraystretch}{1.15}
\begin{tabular}{lccc}
\hline
\multicolumn{4}{c}{Efficiency (normalized by Greedy)} \\
\hline
Method & Token cost ($\times$) & Time ($\times$) & Main hyperparams \\
\hline
\multicolumn{4}{c}{Gemma-3-4b-it (4B)} \\
\hline
Greedy & 1.00 & 1.00 & -- \\
Beam Search & 3.32 & 1.88 & beam=$5$ \\
Self-Consistency & 2.90 & 2.80 & $n=5$ \\
Steering & 0.93 & 0.99 & $\alpha=0.5$ \\
TTPO & 1.26 & 1.12 & -- \\
Branching & 1.18 & 1.45 & $K{=}20, L{=}20$ \\
\hline\hline
\multicolumn{4}{c}{Phi-4-mini-instruct (4B)} \\
\hline
Greedy & 1.00 & 1.00 & -- \\
Beam Search & 5.61 & 1.75 & beam=$5$ \\
Self-Consistency & 4.37 & 4.09 & $n=5$ \\
Steering & 0.89 & 1.00 & $\alpha=0.5$ \\
TTPO & 0.89 & 0.93 & -- \\
Branching & 1.02 & 1.53 & $K{=}20, L{=}20$ \\
\hline
\end{tabular}
\caption{Efficiency for 4B models. Token-cost counts the total number of next-token forward steps (including lookahead branches) normalized by greedy. Time is measured wall-clock latency normalized by greedy under the same hardware and decoding setup.}
\label{tab:eff_cost_4b_combined}
\end{table*}

\begin{center}
\subsection{BIG-Bench Hard (deductive subset)}
\begin{promptlisting}[BIG-Bench Hard (deductive subset) Prompt]
|\textcolor{blue}{[System Instruction]}|
You are an expert in logical reasoning and reading comprehension. Your task solve questions.

Follow these steps strictly:
1. Reasoning step by step.
2. Select the answer in the format '/boxed{ANSWER}'. for example, if the answer is option A, the output should be '/boxed{A}'

|\textcolor{blue}{[User Prompt]}|
# Context:
|\textcolor{red}{[context]}|

# Question:
|\textcolor{red}{[question]}|

# Choices:
|\textcolor{red}{[choices]}|

Think step by step.

\end{promptlisting}
\captionof{figure}{Big Bench Hard (deductive subset) prompt template}
\label{fig:big_bench_hard_prompt}
\end{center}

\section{Efficiency Comparison}
\label{sec:efficiency}

We report an efficiency comparison between decoding baselines and our connective centric methods.
All efficiency numbers are measured on the \textit{ZebraLogic test split} with Gemma-3-4b-it model and are \textit{normalized by greedy decoding} for each model (Greedy $=1.00$), so values $>1$ indicate additional compute/latency relative to greedy.

\paragraph{Token cost ($\times$ Greedy).}
Token cost counts the total number of next token forward steps required by each method, aggregated over the test set and normalized by greedy. This includes any extra forward passes introduced by the decoding algorithm, e.g., multiple sampled trajectories for \textit{Self-Consistency} ($n=5$), multiple hypotheses maintained for \textit{Beam Search} (beam size as in Table~\ref{tab:main_results}), and additional lookahead generation for \textit{Branching} (triggered at connective pivots with hyperparameters $K$ and lookahead length $L$). In contrast, Steering and TTPO preserve single trajectory greedy decoding at inference time; their token cost deviations from 1.0 mainly reflect changes in the generated length, not additional search.

\paragraph{Time ($\times$ Greedy).}
Time is measured as \textit{wall-clock latency} from the start of generation to termination (EOS or max length), also aggregated over the ZebraLogic test set and normalized by greedy.

\paragraph{Hyperparameters.}
For reference, we summarize the main decoding hyperparameters in Table~\ref{tab:eff_cost_4b_combined}: beam size for beam search, $n$ for self-consistency, $\alpha$ for steering, and $(K,L)$ for branching. These knobs directly control the amount of additional inference time computation for search based methods, while TTPO modifies model behavior through training and does not introduce inference time search.
\end{document}